# Explainable Multi-class Classification of the CAMH COVID-19 Mental Health Data

YuanZheng Hu
EECS, University of Ottawa

*bhu078@uottawa.ca*

Marina Sokolova
IBDA@Dalhousie University and
University of Ottawa

*sokolova@uottawa.ca*

## Abstract

Application of Machine Learning algorithms to the medical domain is an emerging trend that helps to advance medical knowledge.  At the same time, there is a significant a lack of explainable studies that promote informed, transparent, and interpretable use of Machine Learning algorithms.   In this paper, we present explainable multi-class classification of the Covid-19 mental health data.  In Machine Learning study, we aim to find the potential factors to influence a person's mental health during the Covid-19 pandemic. We found that Random Forest (RF) and Gradient Boosting (GB) have scored the highest accuracy of 68.08% and 68.19% respectively, with LIME prediction accuracy 65.5% for RF and 61.8% for GB.  We then compare a Post-hoc system (Local Interpretable Model-Agnostic Explanations, or LIME) and an Ante-hoc system (Gini Importance) in their ability to explain the obtained Machine Learning results.  To the best of these authors' knowledge, our study is the first explainable Machine Learning study of the mental health data collected during Covid-19 pandemics.

## Introduction

Machine Learning algorithms applied to the medical domain is an emerging trend that helps to advance medical studies.  Machine Learning (ML) and Deep Learning (DL) algorithms are often deployed to analyse large and diverse data sets when a timely response is essential. Classifications of medical images in respect to the COVID-19 diagnosis (Mohamadou et al, 2020), the Covid-19 forecasting model by Google Cloud and Harvard Global Health Institute help the frontline medicine.  At the same time, reported classification accuracy and predicted infections, hospitalizations, expected deaths tell only a part of the story if the studies use a black box approach where the algorithms' internal factors are treated as either unknown or beyond interpretation.   The black box approach impends successful implementation and reproducibility of ML and DL studies that depend on a detailed and systematic analysis of the models, learning functions involved, meta-parameter influence on the obtained results, among others.

In this work, we demonstrate how post-hoc and ante-hoc explanations enrich ML studies.  Multi-class classification of the Covid-19 Mental Health National Survey[1] data serves as the ML base of our work. The dataset is extracted from a series of six surveys conducted in Canada during May – December 2020.  The surveys aimed to investigate mental health during the pandemic in Canada.    Our goal is to find out the potential factors to influence a person's mental health during the Covid-19 pandemic.

---

[1] https://www.camh.ca/en/health-info/mental-health-and-covid-19/covid-19-national-survey

We classify the surveys' participants into one of the six categories, where each category corresponds to a survey. The survey questions are the data features; the participant answers are feature values. We use six algorithms (Gradient Boosting, Random Forest, Decision Tree, SVM, Logistic Regression, Naïve Bayes). We apply a post-hoc system LIME (Holzinger et al, 2017) to explain the predictions of the six Machine Learning algorithms. After we train our dataset using the six models (Gradient Boosting, Random Forest, Decision Tree, SVM, Logistic Regression, Naïve Bayes), we use an ante-hoc system Gini Importance to analyze two Machine Learning models that achieve the best results (Gradient Boosting ad Random Forest).

We present a comprehensive analysis of the LIME prediction accuracy for Random Forest and Gradient Boosting and compute LIME probability estimates for the top most predictions for the six ML classifiers. We compare LIME and Gini Importance results by using the explainability fact sheet (Sokol and Flach, 2020). The fact sheet lists functional requirements, operational requirements, usability, safety, and validation as key aspects of explainability. We show that LIME and Gini Importance are similar in operational requirements and differ in functional requirements.

Our explanation results show that consumption of alcohol and use of cannabis have a strong positive impact on determining the periods of the pandemic. This result helps to get insights into the general public's mental health during the Covid-19. At the same time, it delivers an important information to decision makers about usage of the recreational drugs in times of crises.

To the best of these authors' knowledge, our study is the first explainable Machine Learning study of mental health data collected during Covid-19 pandemics. Our study fills the void in post-hoc and ante-hoc explanations of multi-class classification of mental health data and comparison of post-hoc and ante-hoc explanation results.

## Explainable Machine Learning

Explainable Machine Learning examines the results given by Machine Learning models with the aim to justify decisions, to enhance control, to improve models, and to discover new knowledge (Adadi and Berrada, 2018). Model's explanation, outcome explanation, model inspection, and transparent model design are essential elements of explainable Machine Learning (Guidotti et al, 2018). It has been shown that the explanation systems improve human prediction of algorithms' behaviour on new inputs (Hase and Bansal, 2020). Whereas explainable ML emerges in business and scientific studies (Roscher et al, 2020), a medical domain remains significantly underserved (Holzinger et al, 2017), especially in multi-class classification tasks. Further we list explainable ML studies that worked with medical data sets.

Risk predictions of Covid-19 has been studied by Casiraghi et.al. (2020). The authors use CT and CXR images, as well as radiological, clinical and laboratory data. They applied Random Forest along with Gini Importance to compute the feature relevance of Covid-19. The authors used *mean decrease in accuracy* for Gini Importance; the obtained results show C-Reactive Protein (CRP) being attributed with the most positive correlation to the risk of Covid-19 together with patient's age. We, au contraire, use *mean decrease in node impurity*, to conduct our study.

Yoo et.al. (2020) used the SHAP technique to explain the XGBoost model when classifying the type of laser refractive surgery that is suitable for the patient (i.e., LASEK, LASIK, SMILE, and contraindication to corneal laser surgery) based on patient's data (i.e., Ocular measurements, questionnaires). Based on the SHAP technique, they found patient's anticipated surgery type is the most influential factor when deciding the type of surgery. The SHAP technique falls into the same category as LIME – post-hoc system, the technique based on the idea of Shapley value – an average value of the marginal contribution in all permutations of the features. However, there is a critical difference: LIME is a local interpretable explanation model, whereas SHAP is a global interpretable explanation model. Based on a set of experiments by Ignatiev (2020), they obtained results similar to Yoo et al, but on average LIME outperforms SHAP in terms of redundancy and correctness in five datasets. Contrary to comparing post-hoc systems, our current work compares post-hoc and ante-hoc explanation models.

In our previous explainability research, we used a white-box explanation to study the hyperparameter impact on multi-class accuracy of Machine Learning models (Hu and Sokolova, 2020). We worked with six models (i.e., RF, GB, LR, DT, SVM, Naïve Bayes) and performed three-class classification of a benchmark set Diabetes 130-US hospitals for years 1999-2008[2]. Our current work continues explainability studies of multi-class classification, expanding the previous work with the analysis of post-hoc and ante-hoc explanation models.

In the current work, we obtained the best classification accuracy with Random Forest (RF) and Gradient Boosting (GB). Those algorithms are successfully used in classification of medical data sets. RF achieved 99.82% and 99.7% accuracy in binary classification of benchmark sets, Wisconsin Breast Cancer Diagnosis Dataset (699 instances) and Wisconsin Breast Prognostic Cancer Dataset (198 instances) (Nguyen et.al.,2013). In a more recent study, Alam et.al (2019) have applied RF to classify 10 medical benchmark data sets from the same UCI repository: Wisconsin Breast Cancer – 699 instances, Pima Indian Diabetes – 768, Bupa - 345, Hepatitis – 155, Heart-Statlog – 270, SpectF – 267, SaHeart – 462, PlanningRelax – 182, Parkinsons – 195, and Hepatocellular Carcinoma – 165 instances. RF achieved the best accuracy of approx. 97% in binary classification of Wisconsin Breast Cancer. Note that such a high accuracy has been obtained on small data sets in binary classification setting. We, on the other hand, classify 6,021 instances in a six-class classification scheme, with a naïve accuracy baseline (approx. 16%) being significantly lower than in the tasks mentioned above.

Gradient Boosting has been applied in clinical medicine, where it outperformed Logistic Regression in a simulated linear regression task by using a simulated dataset with 98% accuracy (Zhang et.al., 2019). Our study, in contrary, is applied to multi-class classification of data gathered in a mental health research. Gradient Boosting has also been used in multi-class classification of data related to personal well-being (Rahman et.al, 2020). The authors used several boosting strategies (i.e., XGB, LGBM, GB, CB, and AdaBoost) to perform a multiclass classification task on daily activities (i.e., Walk, Upstairs, Downstairs, Sit, Stand, and Lie). The study used data generated by wearable sensors; the result showed that GB and ADA achieved the best accuracy (93.9%) among all the boosting-based algorithms. Our work, on the other hand, uses the data gathered from 6 surveys open to the general public.

## Explanation Systems:  LIME and Gini Importance

Currently, explanation systems employ either post-hoc or ante-hoc principles. Post-hoc systems first treat ML models as a black-box and give specific explanation for instances after the model training. Ante-hoc systems, conversely, embed explainability into the model during the training time, and thus build an explanation model using a white-box (or open-box) approach, as the embedded algorithms in the models are considered as open-source techniques (Kojarski et.al 2006, Filip Anderson, 2016). In this work, we use LIME, a post-hoc system, and Gini importance, an ante-hoc system.

---

[2] https: //archive.ics.uci.edu/ml/datasets/diabetes+130-us+hospitals+for+years+1999-2008

## Local Interpretable Model-Agnostic Explanations

Local Interpretable Model-Agnostic Explanations (LIME) is a Post-hoc system coined by Ribeiro et.al. (2016). LIME aims to produce an easily interpretable explanation for the non-professionals to understand predictions of Machine Learning models. The LIME name suggests that the algorithm explains an instance of the test samples locally with respect to the features it has. This technique can apply to any model, regardless of the internal mechanism of the model.

The basic idea behind LIME can be explained by Figure 1, adapted from Ribeiro et.al. (2016). The red cross lies in the pink area stands for the instance we picked to explain, and the pink/blue background stands for the classification problem we are dealing with. LIME will tune the values of the features from the picked instance and generate new samples based on the proximity to the instance being picked. At last, LIME will optimize the dotted line based on all generated samples and give a local interpretable explanation of the instance being picked.

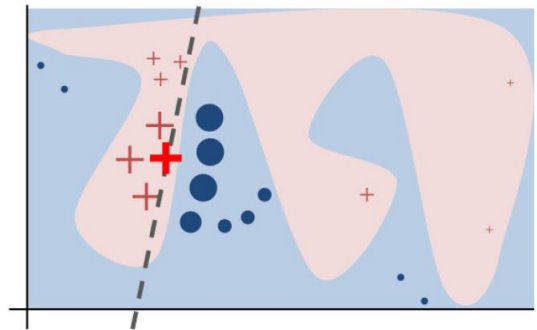

*Figure 1. LIME (by Ribeiro et.al., 2018)*

There are three important characteristics for LIME, namely, interpretability, local fidelity, and model agnostic. For interpretability, the model should provide explanation easily understandable by users of ML systems, who may not be professionally trained in Machine Learning.

For a local fidelity, the explanation produced by the model should be locally faithful to the instance being picked. For a model agnostic, the explanation model should be able to explain any model. A formulated explanation can be found in the formula below, where Equation 1 represents the LIME explanation model.

*Equation 1: the LIME model*

$$\xi(x) = \underset{g \in G}{\operatorname{argmin}} \; \mathcal{L}(f, g, \pi_x) + \Omega(g)$$

To retain both local fidelity and interpretability, the formula consists of two parts. The first term is a measure of the unfaithfulness of $g$ (potential explanation model) in approximating $f$ (original predictor) where $g$ is locality defined by $Pi(x)$ (proximity measure between the sample and the picked instance).

The second term stands for the complexity of the explanation of the LIME model. LIME aims to minimize unfaithfulness and lower complexity of the explanation model.

In a high-quality human users' test, LIME outperformed Anchor, Decision Boundary, a Prototype model, and a Composite approach when those systems were assessed in improving human ability to predict a ML classification model behaviour (Hase and Bansal, 2020).  Users' prediction accuracy of the ML classification increased 11.25% (significant at a level of $p < .05$, CI = 95%) after they been given LIME explanations.   The other tested systems yield 5.01%, 1.68%, 5.27%, and 0.33% accuracy increase, respectively.  The experiments were conducted on classification of the Adult data set[3].

However, LIME's dependence on heuristics and local fidelity property can cause certain drawbacks. Heuristic explanations are unable to catch all the properties of the underlying ML models, especially when counterexamples exist (Ignatiev, 2020).  On entire instance space, i.e., in presence of counterexamples, explanations of LIME and other heuristic models (Anchor, SHARP) were mostly incorrect in 4 out 5 datasets.   At the same time, the counterexample disposition also influences LIME's performance.  LIME, and Anchor, improved users' ML accuracy prediction in a counterfactual test, when the users were asked to predict ML model behaviour on perturbations of the previously given data entries (Hase and Bansal, 2020).

Thus, the LIME performance may be adversely influenced by a high number of counterexamples.  To reduce the negative impact of counterexamples, we recommend an appropriate choice of classification evaluation measures as the classification criteria, e.g., Recall in case of positive counterexamples. Refer to Sokolova and Lapalme (2009), Sec 4., for detailed discussion of reliability of the classification performance measures in the presence of counterexamples.

### Impurity-based Feature Importance

Gini Importance, an Impurity-based approach, is an ante-hoc system (Holzinger et al, 2017).  In difference with LIME and other post-hoc methods, the impurity-based approach is used internally by the Machine Learning models. Explanation provided by such models reflect ML model's internal mechanism, hence are more reliable than explanations produced by post-hoc systems.

There are three common approaches of measuring impurity: Gini importance, information entropy, and misclassification rate.  Information Entropy is a measurement of the disorder of information (Shannon, 1948). Decision tree-based models adapt this idea and uses entropy-based criterion to find the largest information gain in each feature when splitting and aims to decrease the information entropy of the entire tree. Similarly, the goal of misclassification-based criterion is to decrease the misclassification rate by picking the feature results in lowest misclassification rate as the splitting node (Badulescu, 2007). In a popular Scikit Learn[4], the feature importance of an ensemble model is measured by Gini Importance.

Scikit-Learn implements Gini Importance by using results of Breiman et al (1984). It defines the Gini Importance as the total decrease in the node impurity averaged over all trees of the ensemble model. In our research, Gradient Boosting, Random Forest, Decision Tree (GB, RF and DT respectively) are tree-based ensemble algorithms. They all use Gini Importance in their implementation.

---

[3] https://archive.ics.uci.edu/ml/datasets/Adult
[4] https://scikit-learn.org/

Equation 2 reports the formula for the definition of Gini Importance. It is the probability of misclassification on a random instance, where the $p_{mk}$ is defined as the proportion of class k in node m. The goal of the tree-based models is to find the feature with high Gini gain to split to decrease the Gini Impurity.

*Equation 2: Gini Impurity*

$$H(Q_m) = \sum_k p_{mk}(1 - p_{mk})$$

Equations 3 and 4 illustrate how Gini Impurity is used for calculating feature importance. Equation 4 assumes a binary decision tree, and the node impurity value is subtracted by the left impurity value and the right impurity value where both values are weighted by the number of training samples that reached the node. On the right-hand side, the formula shows the importance value of feature *i*, it is calculated by the sum of all node impurity value splits on that feature then divided the sum of all nodes' impurity value. Since we are using an ensemble tree model, the feature importance of an ensemble model can be calculated by further normalized the value within the range of 0 to 1, and then averaged by the number of trees used in the model.

*Equation 3. Node importance*

$$ni_j = w_j C_j - w_{left(j)} C_{left(j)} - w_{right(j)} C_{right(j)}$$

*Equation 4. Feature importance*

$$fi_i = \frac{\sum_{j:\text{node } j \text{ splits on feature } i} ni_j}{\sum_{k \in \text{all nodes}} ni_k}$$

Gini Importance served as the criterion to split into attributes of RF and DT in the most used ML packages (i.e., scikit-learn, Weka). Information Gain is another splitting algorithm used by tree-based models. In empirical evaluations, Gini Importance and Information Gain yield no significant difference in Accuracy, Recall, and Precision (Tangirala, 2020). However, as pointed out in (Breiman et.al, 1984; Strobl et.al, 2007), both Gini Importance and Information Gain are biased towards multivalued attributes because the level (the number of categories) of multivalued attributes is directly associated the expected value of Gini Importance. Such bias does not affect our study since we convert all categorical features to numerical values.

## The Covid-19 Mental Health Dataset

A Kaiser Family Foundation (KFF) poll, conducted in US amid the COVID-19 pandemics March – July 2020, reported a variety of negative mental health factors among respondents: sleeping (36%), overeating (32%), headache or stomachache (18%), bad temper (18%), and an increased usage of drug and alcohol (12%) (Hamel et al, 2020). Our research focuses on investigating the influential factors of mental health during the COVID-19 pandemic. The CAMH Coronavirus Mental health dataset[5] is a part of an ongoing study by the Centre for Addiction and Mental Health (CAMH). The study, *Examining the Impact of COVID-19 on Mental Health and Substance Use among Canadians*, aims to understand the mental health and substance use impacts of COVID-19.[6] The dataset has been built from the results of six national (Canadian) survey reports conducted during different waves of the COVID-19 pandemic.[7] All participants were English-speaking Canadians ages 18 and older. They accessed the surveys through an online portal Asking Canadians.[8] The portal does not provide immediate monetary rewards. Instead, the participants can either automatically earn reward points of a few selected companies or enter contests and draws.

The surveys were conducted on May 8 to 12, 2020, (1,005 respondents); May 29 to June 1, 2020, (1,002 respondents); June 19 to 23, 2020 (1,005 respondents); July 10 to 14, 2020 (1,003 respondents); September 18 to 22, 2020 (1,003 respondents), and November 27 to December 1, 2020 (1,003 respondents). Our data set has 6,021 instances in total. We have accessed the data on *delvinia.com.* Each survey has 29 questions that are used to test the mental health – The first three questions seek general information, i.e., age, gender, and location of the participants. Questions 4 - 8 ask about the impact of the pandemic, e.g., salary, COVID test positives. Question 9 - 15 focus on the mental health of the participants, e.g., anxiety, nervousness. Question 15 - 23 seek recent changes in the habits of the participants, e.g., alcohol, cannabis. The last six questions involve private matters such as income, number of family members, etc. We list the questions and their options detail in Appendix 1.

The dataset we downloaded is in the form of csv. In the original dataset, all questions with numerical options are labeled with their actual value. For example, Question 23 asks the number of family member, then the feature for this question is labeled by the actual value of the number of participant's family member. For questions without numerical options, the features are one-hot encoded. We list the details of the features type of dataset in Table 1.

| Encoding method | Questions/Features |
| --- | --- |
| Numerical values | Q1,2,3,5,6,7,8,9,10,11,12,13,14,15,16,17,18,19,20,21,22,23,25,26,27,28,29 Respid, language, agreement, hage, gender, hIncome, hChild, hHousehold, hGender, hRegeion, hWave |
| One-hot encoding | Q4,24 |
| String | Status |

*Table 1, Question feature types in original dataset*

Note that Q1-29 represent answers by the participants. The remaining questions represent participants' metadata that was not shown to the participants. However, metadata is a part of dataset. We use it in Machine Learning experiments.

---

[5] htttps://www.delvinia.com/camh-coronavirus-mental-health/
[6] https://www.camh.ca/en/health-info/mental-health-and-covid-19/covid-19-national-survey
[7] https:// www.camh.ca/en/camh-news-and-stories/anxiety-patterns-in-canadians-mirror-progression-of-pandemic
[8] https://www.askingcanadians.com/communities/default.aspx?p=p430686807&dlvl=9

## Data Preprocessing

Machine Learning requires features to be consistent when representing the data items. The COVID-19 Mental Health data set does not support such consistency for all the features. Although the survey questions stay the same for the duration of the study, the response options for Q4 changes, to allow for more diverse answers as the pandemic progresses. Q4 asks if you or someone you know has been tested positive for Covid-19; in the May 8-12 survey, "you" and "someone" are put together as one response option, but in the Nov 27 – Dec 1 survey, "you" and "someone" are two response options. Q23CP appears only in surveys 5 and 6. Our solutions are listed in Table 2.

| Question | Options Difference | | Solution |
|---|---|---|---|
| Q4 | A: I have test positive … B: Someone have test positive … Wave 1,2,3,4 | A: I/Someone have test positive .. Wave 5,6 | Combine separate options in surveys 1,2,3,4 |
| Q23CP | No such questions surveys 1,2,3,4 | Extra questions surveys 5,6 | Remove Q23CP from the dataset |

*Table 2, Questions with different answer between different waves*

## Data Construction

We concatenate six data frames extracted from the six surveys' reports. Note that the number of respondents is balanced for the reports, thus the resulting data is balanced with respect to the survey representation. We observe that some questions were not shown to the participants, i.e., features without question label in Table 1. We remove those features since they are not directly answered by the respondents; thus, they are useful for the purpose of our research.

In addition to the data consistency, we consider features like survey status, survey ID, consent agreement and language options to be non-essential and remove them from the data. Lastly, we need to handle the missing data in the original dataset, we plot the missing data in Fig 2.

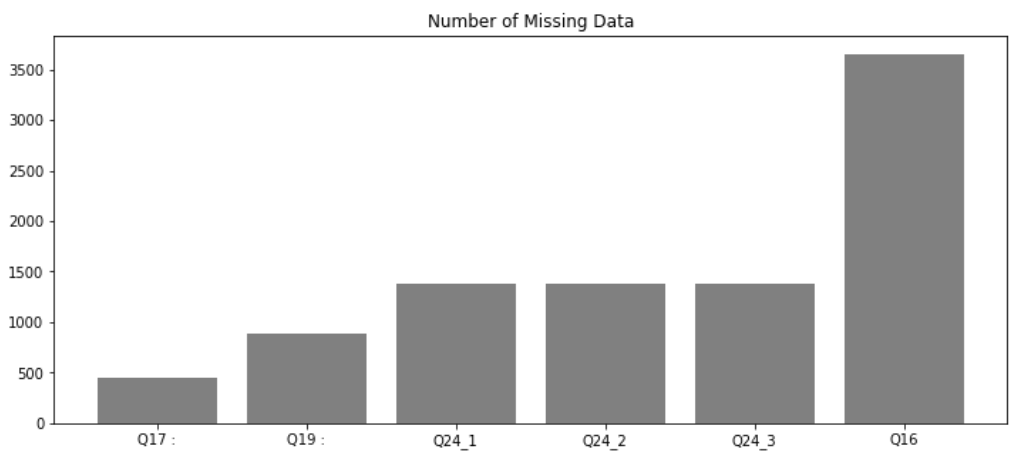

*Figure 2, Number of missing data in original dataset*

We can see that the data for Q16 is highly missing, thus we drop this feature from the further study. For the remaining five features, we use the mean value to smooth the missing data. For all the other features, we consider them well preprocessed and keep them as in the original dataset.

# Machine Learning Experiments

We multi-classify the COVID – 19 Mental health data set into one of the six surveys. We seek ML models that achieve the best classification accuracy on the COVID-19 Mental Health data. (Further on, we use LIME and Gini Importance to examine those selected models.) We have used Gradient Boosting, Random Forest (RF), Decision Tree, SVM, Logistic Regression, and Naïve Bayes. We use the Scikit Learn default settings, then perform parameter tuning to improve the classification accuracy. We conduct parameter tuning as reported in (Hu and Sokolova, 2020). Fig 3 visualizes our classification results before and after the parameter tuning. Table 3 reports the best results for the top three models.

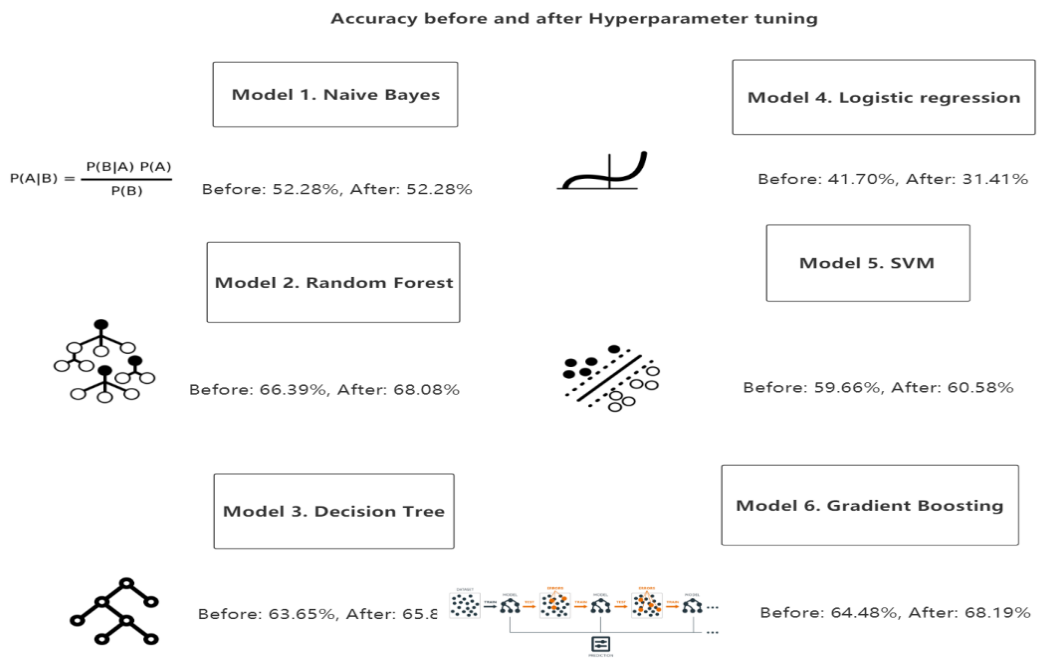

*Figure 3, Machine Learning results obtained by 10-fold cross-validation*

| Model | Best Config | Accuracy Results | Result Improvement |
|---|---|---|---|
| Gradient Boosting | lr: 0.1<br>max_depth: 1<br>n_estimator: 150 | 68.19%, | Significant, 4%↑ |
| Random Forest | max_depth: 6<br>max_feature:61<br>min_sample_split:4<br>n_estimator:500 | 68.08% | Insignificant, 1.2% ↑ |
| Decision Tree | *max_depth:20*<br>*max_feature: 61*<br>*min_sample_leaf: 3*<br>*min_sample_split: 3* | *65.80%* | *Significant, 2.15%↑* |

*Table 3, Model evaluation after parameter tuning*

To summarize, GB and RF achieved the best results among the six models, with accuracy of 68.19% and 68.08% respectively. Gradient Boosting has the largest accuracy increase after parameter tuning with around 4% of the increase in accuracy. This corresponds to the multiclass classification results reported in Hu and Sokolova (2020). A more detailed evaluation can be found in Appendix 2.

## LIME application

We re-trained our models on a stratified dataset using configuration obtained by the parameter tuning. For each cohort, we use 80% of the data as training set and the remaining 20% of data as test set, i.e., 800 instances for training and 200 instances for the test. Note that since LIME takes the prediction function of the models as input, so the accuracy of LIME prediction is the same as the accuracy of the models. For the LIME application, we picked all instances from the test dataset and trying to find out the performance of the LIME prediction. (See Appendix 3 for a case study). We report the total LIME prediction accuracy for RF and GB in Table 4.

The results show that LIME predictions align for both classifiers: they were the lowest for the survey # 4 classification (40.0% for RF and 41.0% for GB) and the highest for survey # 2 classification (94.0% for RF and 99.0% for GB).

*Table 4: LIME prediction accuracy for RF and GB.*

|  | Radom Forest | | | Gradient Boosting | | |
| --- | --- | --- | --- | --- | --- | --- |
|  | Correctly predicted | incorrectly | Accuracy | Correct | incorrect | Accuracy |
| Survey #1 | 183 | 18 | 91.0% | 181 | 20 | 90.0% |
| Survey # 2 | 188 | 12 | 94.0% | 198 | 2 | 99.0% |
| Survey # 3 | 114 | 87 | 57.0% | 110 | 91 | 55.0% |
| Survey # 4 | 80 | 121 | 40.0% | 82 | 119 | 41.0% |
| Survey # 5 | 124 | 77 | 62.0% | 128 | 183 | 64.0% |
| Survey # 6 | 100 | 101 | 50.0% | 114 | 87 | 57.0% |
| Total | 789 | 416 | 65.5% | 813 | 502 | 61.8% |

We collect the prediction probabilities of all the test samples to calculate the average probability of the highest prediction for each model. Our results show that Naïve Bayes has the highest probability for the *top most* prediction among all models, 89.92%, when the other models get lower results on their *top most predictions*: RF - 54.04%, GB - 67.57%, LR - 27.23%, DT - 81.11%, and SVM- 61.76%. Thus, although Naïve Bayes gives a low accuracy of classification (i.e., 52.28%), the LIME model has the highest confidence on its top prediction compared to the other models.

To further investigate the potential factors that influence LIME's ML model's prediction for each COVID-19 survey, we take the absolute value of feature importance of all the features on every test instance and add them up, then pick the top five features with the highest feature importance for both models. Tables 5 and 6 display the obtained results. $Q4\_i$ ($i = 1, ..., 6$) means one person close to the survey responder either has tested positive for Covid-19 or at high risk of Covid-19. Q4_7 means seven people and more. Q18 and Q19 relate to use of cannabis.

| Survey | Random Forest | | | | |
|---|---|---|---|---|---|
| | Top1 | Top2 | Top3 | Top4 | Top5 |
| 1 | Q19 | Q4_1 | Q18 | Q16 | Q7 |
| 2 | Q4_1 | Q18 | Q19 | Q17 | Q7 |
| 3 | Q4_1 | Q18 | Q19 | Q4_3 | Q7 |
| 4 | Q4_1 | Q18 | Q19 | Q4_4 | Q4_3 |
| 5 | Q4_1 | Q18 | Q19 | Q4_2 | Q4_5 |
| 6 | Q4_1 | Q18 | Q19 | Q4_2 | Q4_5 |

*Table 5, Top five influential questions of each survey for RF - LIME*

| Survey | Gradient Boosting | | | | |
|---|---|---|---|---|---|
| | Top1 | Top2 | Top3 | Top4 | Top5 |
| 1 | Q19 | Q4_1 | Q7 | Q4_3 | Q7 |
| 2 | Q19 | Q4_1 | Q7 | Q4_3 | Q4_7 |
| 3 | Q4_1 | Q7 | Q4_3 | Q7 | Q4_4 |
| 4 | Q4_1 | Q4_2 | Q7 | Q4_3 | Q4_4 |
| 5 | Q4_1 | Q4_3 | Q4_2 | Q4_4 | Q4_7 |
| 6 | Q4_1 | Q4_3 | Q4_4 | Q4_7 | Q4_2 |

*Table 6, Top five influential questions of each survey for GB – LIME*

Q4_1, Q18 and Q19 are the top 3 most frequent questions when determining the cohort of the pandemic for both RF and GB. We re-trained RF and GB using only these three features. However, the newly obtained multi-classification accuracy deteriorated: for RF – to 62.66%, for GB – to 63.82%.

## Impurity-based Feature Importance Application

Figures 4 and 5 show the feature importance of GB and RF generated by Scikit Learn. As the graphs indicate, Q19, Q4_1, and Q18 are the top 3 important factors for both RF and GB.

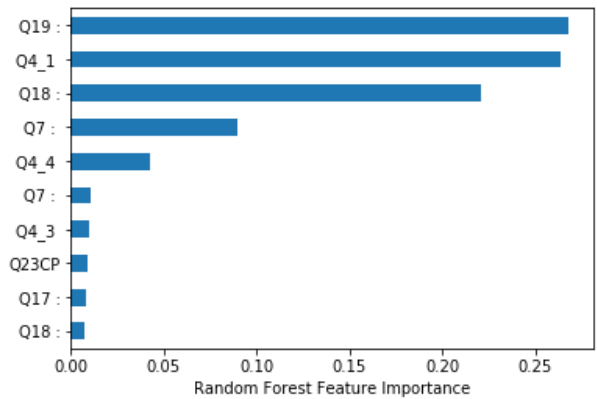
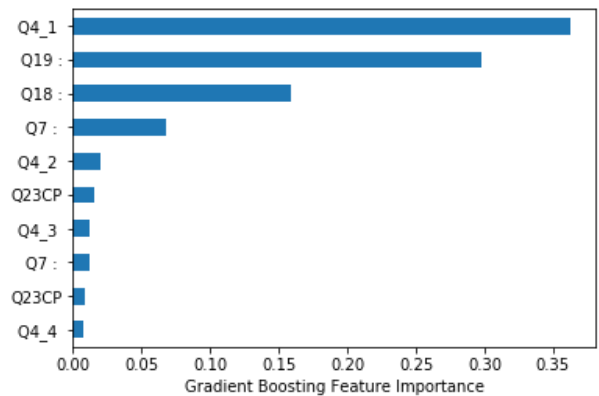

*Figure 4. Ranked feature importance for RF*    *Figure 5. Ranked feature importance for GB*

After we trained the models with top five features, we have 62.32% accuracy for RF and 65.39% accuracy for GB, compare our new result with the results on page 9, we do not have an increase for both models. Compare the results of LIME model and Gini importance, we find that Q4_1, Q18 and Q19, the most frequent positive impact features in LIME prediction are also appear in the top five influential features in both RF and GB predictions.

## Assessment of the explainable models

To compare LIME and Gini Importance systems, we adapted the explainable fact sheet from Sokol and Peter Flach (2019). The fact sheet aims to analyze an explanation model in five dimensions – functional, operational, usability, security, and validation. In Table 6, we compare functional, operational and the soundness in usability characteristics of LIME and Gini importance; the remaining assessment can be found in Appendix 4.

|  | **LIME** | **Gini Importance** |
|---|---|---|
| **Functional Requirements** | | |
| F1: Problem supervision level | Supervised ☑<br>Unsupervised<br>Semi-supervised ☑<br>Reinforcement | Supervised ☑<br>Unsupervised<br>Semi-supervised ☑<br>Reinforcement |
| F2: Problem Type | Classification ☑<br>Regression ☑<br>Clustering | Classification ☑<br>Regression<br>Clustering |
| F3: Explanation Target | Data<br>Models<br>Prediction ☑ | Data<br>Models ☑<br>Prediction ☑ |
| F4: Explanation Breadth/Scope | Local ☑<br>Cohort<br>Global | Local<br>Cohort<br>Global ☑ |
| F5: Computational Complexity | Ω(g)  g is the model complexity | O(fc) f is the number of feature, and c is the number of category of each feature |
| F6: Applicable Model Class | Model-agnostic ☑<br>Model class-specific<br>Model-specific | Model-agnostic<br>Model class-specific<br>Model-specific ☑ |
| F7: Relation to the Predictive System | Ante-hoc<br>Post-hoc ☑<br>(Global) mimic approach | Ante-hoc ☑<br>Post-hoc<br>(Global) mimic approach |
| F8: Compatible Feature Types | Tabular ☑<br>Images (convert to binary) ☑<br>Text (convert to binary) ☑ | Tabular ☑<br>Images (convert to binary) ☑<br>Text (convert to binary) ☑ |
| F9: Caveats and Assumptions |  |  |
| **Operational Requirements** | | |
| O1: Explanation Family | Association between antecedents and consequent ☑<br>Contrasts and differences<br>Casual mechanisms | Association between antecedents and consequent ☑<br>Contrasts and differences<br><br>Casual mechanisms |
| O2: Explanatory Medium | (statistical) summarization<br>Visualization ☑<br>Textualization<br>Formal argumentation<br>Mixture of above | (statistical) summarization<br>Visualization ☑<br>Textualization<br>Formal argumentation<br>Mixture of above |
| O3: System Interaction | Static ☑<br>Interactive | Static ☑<br>Interactive |
| O4: Explanation Domain | Original domain ☑<br>Transformed domain<br>Interpretable data representation ☑ | Original domain ☑<br>Transformed domain<br>Interpretable data representation |
| O5: Data and Model Transparency | Transparent (tabular data) ☑<br>Opaque | Transparent (tabular data) ☑<br>Opaque |
| O6: Explanation Audience | Expert<br>General knowledge ☑<br>Lay audience | Expert<br>General knowledge ☑<br>Lay audience |
| O7: Function of Explanation | Explaining ☑<br>Accountability ☑<br>Fairness | Explaining ☑<br>Accountability ☑<br>Fairness |
| O8: Causality vs. Actionability | Actionable ☑<br>Casual | Actionable ☑<br>Casual |
| O9: Trust vs. Performance | Trust ☑<br>Predictive performance | Trust ☑<br>Predictive performance ☑ |
| O10: Provenance | Predictive model ☑<br>Dataset ☑ | Predictive model ☑<br>Dataset ☑ |
| **Usability Requirement** | | |
| U1: Soundness | $R^2 = 0.56.$ | Not applicable |

*Table7 , Explainability Fact sheet for LIME and Gini Importance*

We find that the operational requirements are similar for the LIME and Gini importance. Operational requirements refer to "how user interact with an explainable system and what is expected from them" (Sokol and Flach, 2019). Other requirements, e.g., functional requirements, algorithmic characteristics, differ between LIME and Gini Importance.

In usability requirements, or properties from the users' point of view, LIME differs from the Gini Importance in three aspects. Those aspects are soundness, completeness, and parsimony. We explained soundness earlier in this section. Completeness varies for the two models as LIME is designed as a local explanation for a specific instance whereas Gini Importance gives global explanation of all features. As for parsimony, LIME supports the top N important features visualization whereas the result of Gini Importance in Scikit-Learn implementation can only achieve parsimony by using external scripts.

LIME and Gini Importance also divaricate in the soundness assessment: whereas post-hoc system (LIME) results can be evaluated in their soundness, this evaluation does not apply to ante-hoc systems (Gini importance) (Sokol and Flach, 2019). For LIME, soundness can be measured by $R^2$ error, a measure of the goodness of a model's explanation:

*Equation 5, R-square formula*

$$R^2 = 1 - \frac{\sum_i(y_i - \hat{y}_i)^2}{\sum_i(y_i - \bar{y}_i)^2}$$

where $y_i$ is the true value, $\hat{y}$ is the predicted value and $\bar{y}$ is the mean of the true value. $R^2$ = 1 represents the best prediction. In our study, $R^2 = 0.56$. We used the function *explainer.score* to compute the value.

For safety and validation requirements – security and effectiveness of explanation model, Gini importance's drawbacks (i.e., bias towards multivalued features) can be easily exploit by the attacker to influence the explanation results. Lastly, validation requirements require further studies for the systems.

## Conclusions and Future Work

In conclusion, in this paper, we have used two Machine Learning explanation methods, Post-hoc system and Ante-hoc, to analyze the factors that influence people during the Covid-19 pandemic. We first feature engineered our dataset and feed it into six commonly seen Machine Learning models to predict the waves of the pandemic by inputting the multiple-choice questions answered by the participants of the dataset. We found that Random Forest and Gradient Boosting are scored the highest accuracy of 68.08% and 68.19% respectively. We have presented a comprehensive analysis of LIME prediction results. LIME prediction accuracy for RF - 65.5%, and for GB - 61.8%. We also have computed LIME average probability of the highest prediction (i.e., the top 1 predictions). That LIME probability was highest for Naïve Bayes – 89.92%; reaching 54.04% for RF and 67.57% for GB.

Then, we have used a post-hoc LIME to explain the factors that impact people during different waves of the pandemic. We have investigated the potential factors that influence LIME's ML model's prediction for each COVID-19 survey. We have found the use of cannabis, alcohol consumption and the number of people diagnosed with COVID have had predominant standings when classifying the data into the survey categories. Next, we have obtained a similar result by using an ante-hoc Gini Importance, an algorithm used internally by GB and RF models. Lastly, we have shown that LIME and Gini Importance are similar in terms of the operational aspect, and different in functional, usability and safety aspects.

For future studies, we hypothesize that a large population sample can reduce a sample bias in determination of participants' mental health conditions. We also hypothesize using multiple labels, e.g., survey #, income, education level, may alter importance of cannabis use and alcohol consumption, as they do not represent ubiquitous behavior among the entire population. As for comparison of post-hoc and ante-hoc explanation models, we can further investigate their soundness, completeness, and context fullness to compare usability of the two different systems.

# Appendix 1 COVID-19 Mental Health Survey by CAMH

| Question ID | Question Label |
|---|---|
| S1 | In which province or territory do you currently live? |
| S2 | To which of the following age groups do you belong? |
| S3 | How do you describe your gender identity? |
| Q4 | (Have you or those close to you (e.g. close relative/friend) tested positive for COVID-19 or are at high risk of COVID-19? (check all that apply) |
| Q5 | How worried are you about the impact of COVID-19 on your personal financial situation? |
| Q6 | How have physical distancing measures due to the COVID-19 pandemic affected your employment situation? (check one only) |

| Q6b | On average how has the number of hours you are working for pay been affected by the COVID-19 pandemic? |
|---|---|
| Q7 | How worried are you that you or someone close to you (close relative or friend) will get ill from COVID-19? |
| Q8x1 | P2W frequency - Feeling nervous anxious or on edge |
| Q8x2 | P2W frequency - Not being able to stop or control worrying |
| Q8x3 | P2W frequency - Worrying too much about different things |
| Q8x4 | P2W frequency - Trouble relaxing |
| Q8x5 | P2W frequency - Being so restless that it's hard to sit still |
| Q8x6 | P2W frequency - Becoming easily annoyed or irritable |
| Q8x7 | P2W frequency - Feeling afraid as if something awful might happen |
| Q15 | During the PAST 7 DAYS on how many days did you drink ALCOHOL? |
| Q16 | On how many of the PAST 7 DAYS did you drink or more drinks on one occasion? A drink means a 341 ml or 12 oz. bottle of beer or cider/cooler (5% alcohol content) a 142 ml or 5 oz. glass of wine (12% alcohol content) or a straight or mixed drink with 43 ml or 1.5 oz. of liquor (40% alcohol content – e.g. rye gin rum). |
| Q17 | In the PAST 7 DAYS did you drink more ALCOHOL about the same or less alcohol overall than you did before the COVID-19 pandemic started? |
| Q18 | During the PAST 7 DAYS on how many days did you use CANNABIS (also known as marijuana hash 'pot')? |
| Q19 | In the PAST 7 DAYS did you use CANNABIS more often about the same or less often overall than you did before the COVID-19 pandemic started? |
| Q20x1 | In the PAST 7 DAYS how often have you felt depressed? |
| Q20x2 | In the PAST 7 DAYS how often have you felt lonely? |
| Q20x3 | In the PAST 7 DAYS how often have you felt hopeful about the future? |
| Q23 | Including yourself how many people are currently living in your household? |
| Q24 | How many children in each of the following categories live in your household?) |
| Q24DK | Prefer not to answer (Q24DK) |
| Q25 | What is the highest level of education you have completed? |
| Q26 | What is your current marital status? |
| Q27 | Which of the following best describes your racial or ethnic group? (Check one only) |
| Q28 | What is the total household income you and other members of your household received in the year ending December 31st 2019 before taxes? Please include income FROM ALL SOURCES such as savings pensions rent and unemployment insurance as well as wages. |
| Q29 | Do you consider yourself to be living in a… |

# Appendix 2 Multi-classification results

Gradient Boosting:   f-1 score/ precision /recall (micro) - 64.48%;  training time - 8.59 sec

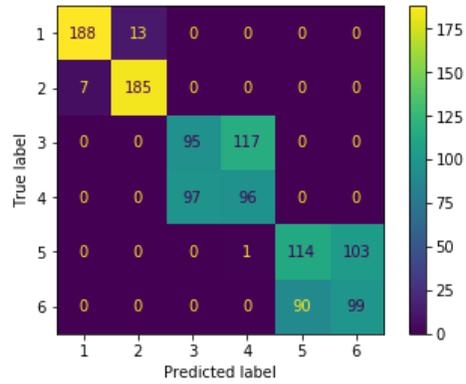

Naïve Bayes: :   f-1 score/ precision /recall (micro) - 52.28%;  training time - 0.013 sec

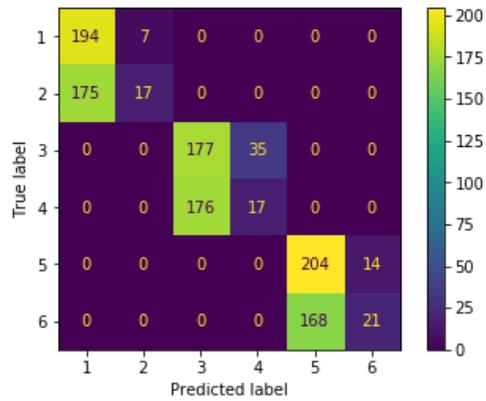

Random Forest: f-1 score/ precision /recall (micro) - 66.39%; training time - 0.814 sec

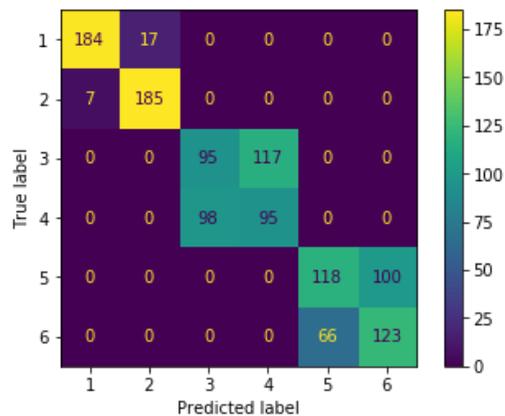

Decision Tree: f-1 score/ precision /recall (micro) - 63.65%; training time - 0.0448 sec

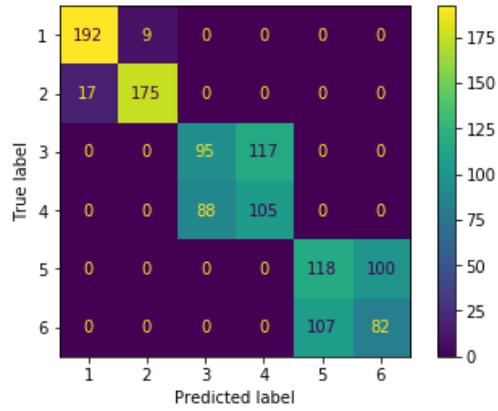

Logistic Regression: f-1 score/ precision /recall (micro) - 31.86%; training time - 7.05 sec

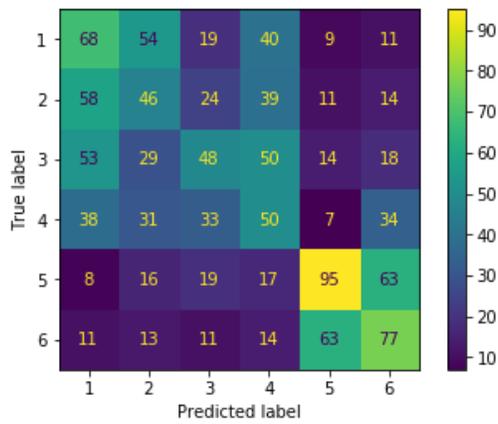

SVM: f-1 score/ precision /recall (micro) - 59.66%; training time - 12.75 sec

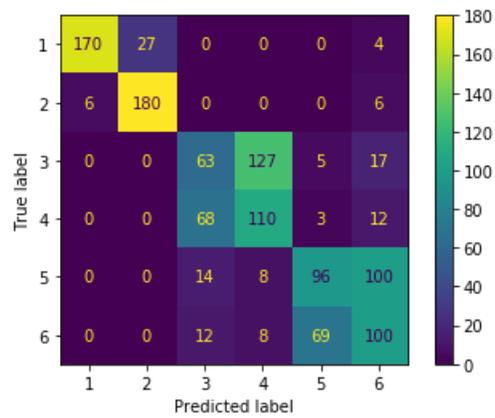

## Appendix 3 LIME Assessment

1) Figures below show explanations from LIME for the classifiers, where the prediction probabilities indicate percent of probability that the prediction is correct. The central plot shows the relative feature importance of the top 5 features in the explanation model, and the left plot shows the actual values for the top 5 features.

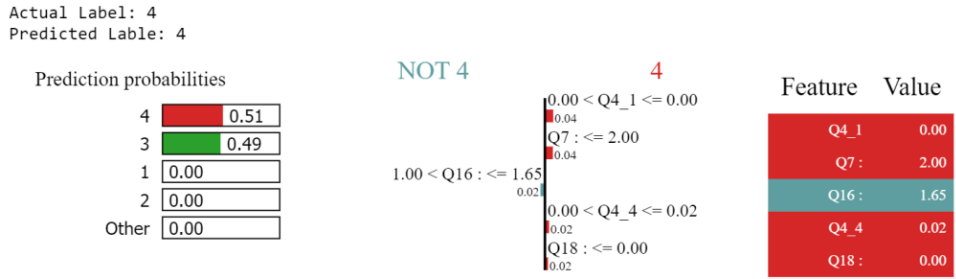

*Figure A3_1. LIME explanation of Random Forest prediction*

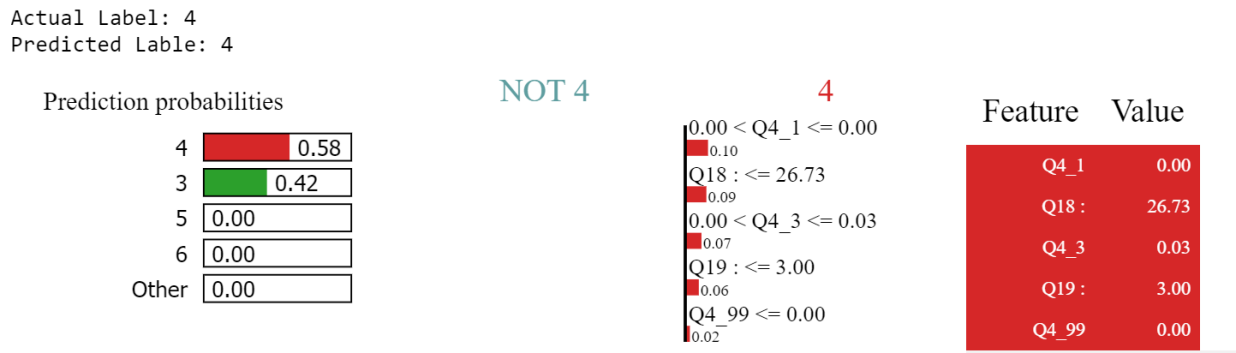

*Figure A3_2. LIME explanation of Gradient Boosting prediction*

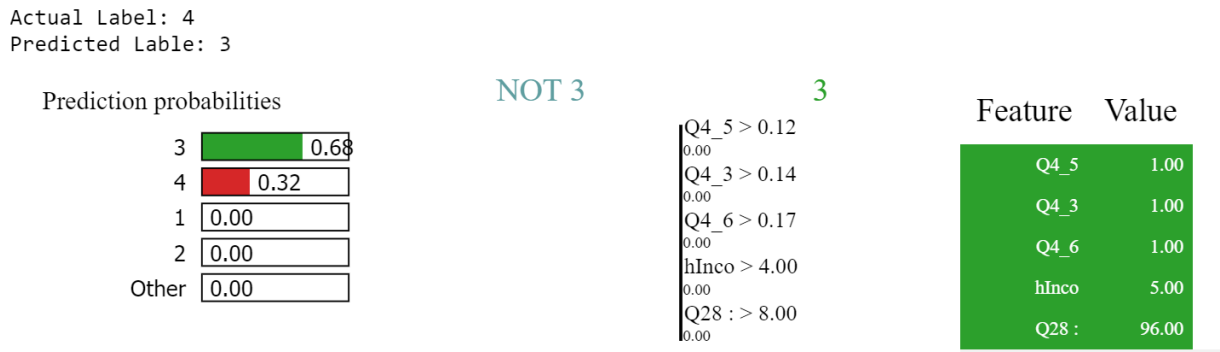

*Figure A3_3. LIME explanation of Naïve Bayes prediction*

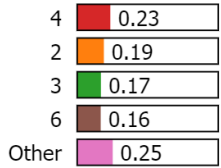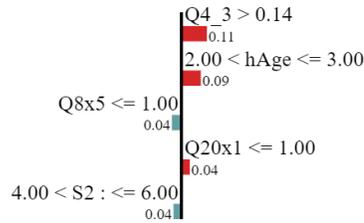

*Figure A3_4. LIME explanation of Logistic regression prediction*

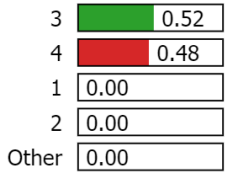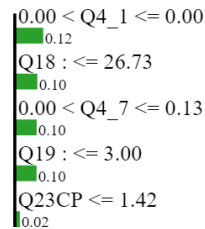

*Figure A3_5. LIME explanation of Decision Tree prediction*

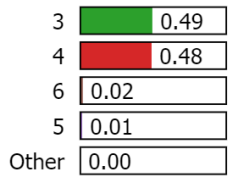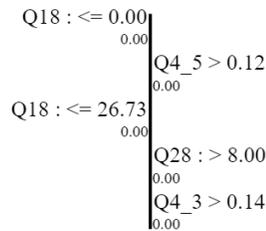

*Figure A3_6. LIME explanation of SVM prediction*

2) The $R^2$ score of LIME is computed as follows: a) the LIME model takes a test dataset with around 1200 instances (20% of the dataset) as input; b) for each test instance, LIME will generate 5000 samples around this instance by default, then a linear model, e.g., Ridge from Scikit-Learn, is used as the potential explanation model $g$ (introduced on page 4) to learn from these 5000 samples using the result from original predictor $f$ (introduced on page 4). We treat the results as the $R^2$ score of the LIME model for the current test instance. Lastly, we compute the $R^2$ score over all 1,200 test instances and take the mean value to estimate the $R^2$ score for the LIME model.

# Appendix 4 The explainable fact sheet for LIME and Gini Importance

| Requirements | LIME | Gini Importance |
|---|---|---|
| **Usability Requirement** | | |
| U2: Completeness | Yes<br>No ☑ | Yes ☑<br>No |
| U3: Contextfullness | Yes<br>No ☑ | Yes<br>No ☑ |
| U4: Interactiveness | Yes<br>No ☑ | Yes<br>No ☑ |
| U5: Actionability | Yes<br>No ☑ | Yes<br>No ☑ |
| U6: Chronology | Yes<br>No ☑ | Yes<br>No ☑ |
| U7: Coherence | Yes<br>No ☑ | Yes<br>No ☑ |
| U8: Novelty | Yes<br>No ☑ | Yes<br>No ☑ |
| U9: Complexity | Yes<br>No ☑ | Yes<br>No ☑ |
| U10: Personalisation | Yes<br>No ☑ | Yes<br>No ☑ |
| U11: Parsimony | Yes ☑<br>No | Yes<br>No ☑ |
| **Safety Requirements** | | |
| S1: Information leakage | Yes<br>No ☑ | Yes ☑<br>No |
| S2: Explanation Misuse | Yes ☑<br>No | Yes ☑<br>No |
| S3: Explanation Invariance | Consistent ☑<br>Inconsistent<br>Stable<br>Unstable ☑ | Consistent ☑<br>Inconsistent<br>Stable ☑<br>Unstable |
| S4: Explanation Quality | Not considered | Not applicable |
| **Validation Requirements** | | |
| V1: User Studies | Section6 of LIME paper | Not applicable |
| V2: Synthetic Experiments | Section 5 of LIME paper | Not applicable |